\documentclass[letterpaper, 10 pt, conference]{ieeeconf}  

\IEEEoverridecommandlockouts                              

\usepackage{amssymb}
\usepackage{amsmath}
\let\labelindent\relax
\usepackage{enumitem}
\usepackage{enumitem}
\usepackage{graphicx}
\usepackage[dvipsnames]{xcolor}
\usepackage{tikz}
\usetikzlibrary{arrows.meta,bending,positioning,calc,math,backgrounds,fit}
\usepackage{booktabs}
\usepackage{algorithm}
\usepackage{algorithmicx}
\usepackage{algpseudocode}
\usepackage{multirow}
\usepackage[separate-uncertainty=true]{siunitx}
\usepackage[font=footnotesize]{caption}
\usepackage[font=footnotesize]{subcaption}

\newcommand{\methodName}{EXPierence replayed, REtrieval augmented,  Specialized VLA}
\newcommand{\methodAbbr}{ExpReS-VLA}

\title{\LARGE \bf
ExpReS-VLA: Specializing Vision-Language-Action Models Through Experience Replay and Retrieval
}

\author{
    \textbf{Shahram Najam Syed}\textsuperscript{1*},
    \textbf{Yatharth Ahuja}\textsuperscript{1*},
    \textbf{Arthur Jakobsson}\textsuperscript{1},
    \textbf{Jeff Ichnowski}\textsuperscript{1} \\
    \textsuperscript{1}Robotics Institute, Carnegie Mellon University, PittsburgFh, USA \\
    *Equal contribution
}

\begin{document}

\maketitle
\thispagestyle{empty}
\pagestyle{empty}

\begin{abstract}%
Vision-Language-Action (VLA) models like OpenVLA demonstrate impressive zero-shot generalization across robotic manipulation tasks but struggle to adapt to specific deployment environments where consistent high performance on a limited set of tasks is more valuable than broad generalization. We present \methodName{} (\methodAbbr{}), a method that enables rapid on-device adaptation of pre-trained VLAs to target domains while preventing catastrophic forgetting through compressed experience replay and retrieval-augmented generation. Our approach maintains a memory-efficient buffer by storing extracted embeddings from OpenVLA's frozen vision backbone, reducing storage requirements by 97\% compared to raw image-action pairs. During deployment, \methodAbbr{} retrieves the $k$ most similar past experiences using cosine similarity to augment training batches, while a prioritized experience replay buffer preserves recently successful trajectories. To leverage failed attempts, we introduce Thresholded Hybrid Contrastive Loss (THCL), enabling the model to learn from both successful and unsuccessful demonstrations collected during deployment. Experiments on the LIBERO simulation benchmark show that \methodAbbr{} improves success rates from 82.6\% to 93.1\% on spatial reasoning tasks and from 61\% to 72.3\% on long-horizon tasks compared to base OpenVLA, with consistent gains across VLA architectures including $\pi_0$ (+3.2 points) and OpenVLA-OFT (+1.7 points). Physical robot experiments across five manipulation tasks demonstrate that our approach achieves 98\% success on both in-distribution and out-of-distribution tasks (with unseen backgrounds and objects), improving from 84.7\% and 32\% respectively for naive fine-tuning. \methodAbbr{} accomplishes this adaptation in 31 seconds using only 12 demonstrations on a single RTX 5090, making it practical for real-world deployment where robots must quickly specialize to their specific operating environment.
\end{abstract}

\section{Introduction}
\label{sec:introduction}

Every deployed robot faces a fundamental paradox: trained on diverse Internet-scale data and robot demonstrations, it must excel at just a handful of tasks in one specific environment. A deployed robot does not require the ability to manipulate all object categories from its 970,000-trajectory training dataset, it requires consistent, high-performance manipulation of the specific objects in its deployment environment.

OpenVLA~\cite{Kim2024}, a 7B-parameter open-source VLA, exemplifies this tension: achieving 70\% success across 29 manipulation tasks, yet struggling to reach the 95\%+ reliability users demand for their specific objects and lighting conditions. This specialization challenge reveals the gap between how we train vision-language-action models, for \emph{broad generalization}, and how we deploy them, for \emph{consistent specialization} in constrained environments.

This specialization challenge manifests as domain shift, subtle differences in lighting, object textures, or spatial layouts that degrade zero-shot performance from acceptable to unusable. While fine-tuning can adapt to specific environments, it can suffer from catastrophic forgetting~\cite{Kirkpatrick2017}, where learning new tasks erases previously acquired skills. Existing solutions either require extensive computational resources (full model fine-tuning on GPUs clusters~\cite{Kim2024}) or fail to leverage failed demonstrations that naturally occur during deployment. Moreover, current approaches treat adaptation as an offline process, incompatible with robots that must improve through daily interaction.

We present \textbf{\methodName{} (\methodAbbr{})}, a method that makes catastrophic forgetting of previously run tasks structurally impossible through frozen encoders and persistent memory buffers, enabling rapid on-device adaptation of pre-trained VLAs. Our key insight is that successful domain adaptation can benefit three complementary mechanisms: (1) compressed memory to efficiently store experiences, (2) retrieval-augmented generation to leverage relevant past experiences, and (3) contrastive learning to explicitly avoid past \emph{failures}. By combining these mechanisms, \methodAbbr{} transforms OpenVLA from a generalist that works adequately everywhere into a specialist that excels in its deployment environment.

\begin{figure}[t]
    \centering
    \includegraphics[width=\linewidth]{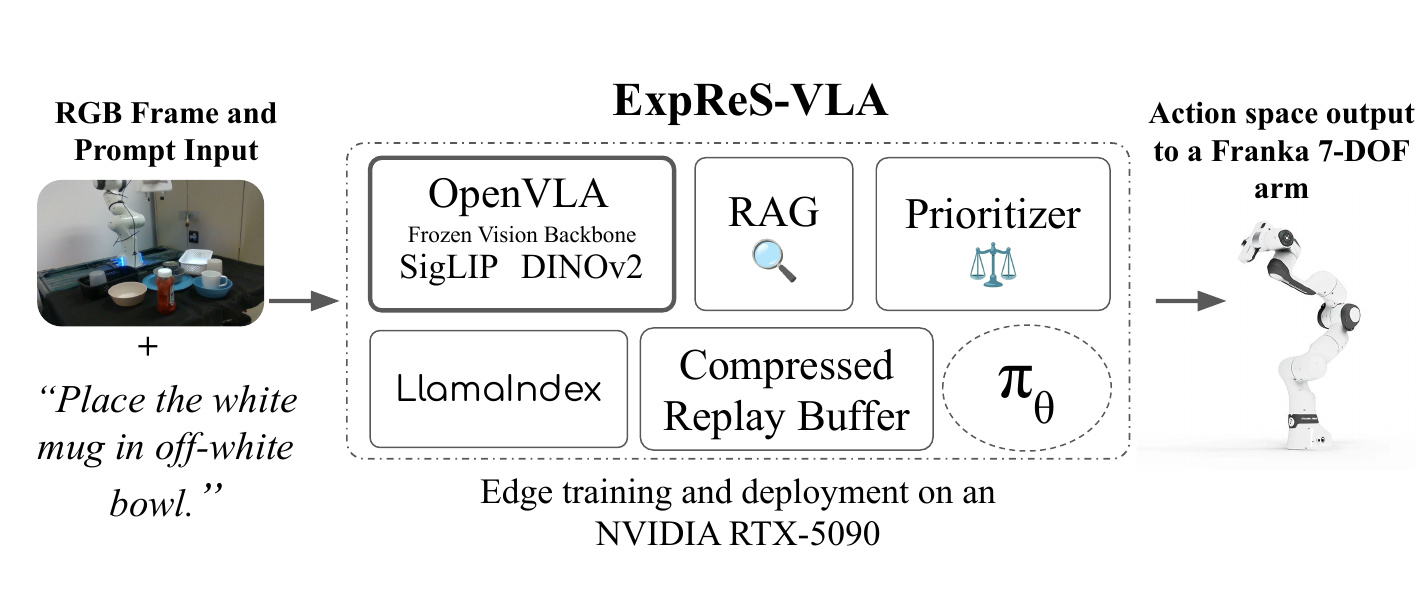}
    \caption{\methodAbbr{} takes in the RGB and prompt input, treats it with encoding, then passes that encoding to the buffer, used for retrivals and prioritizing the learning data for the policy. All of this runs on a single edge device, RTX 5090, and is optimized for performance on the Franka 7-DOF arm.}
    \label{fig:figure-1}
\end{figure}

\methodAbbr{} addresses three critical challenges in practical VLA deployment. First, we achieve 97\% storage reduction for experience replay by storing vision encoder embeddings instead of raw images, enabling efficient memory management for continual learning. Second, we accelerate convergence through retrieval-augmented training that injects contextually similar past experiences into each batch. Third, we introduce Thresholded Hybrid Contrastive Loss (THCL), which adaptively switches between triplet~\cite{Schroff2015} and InfoNCE~\cite{Oord2018} objectives based on failure complexity, transforming unsuccessful attempts into learning signals.

We evaluate \methodAbbr{} in simulation and real-world experiments. On LIBERO simulation benchmarks, \methodAbbr{} achieves 92.4\% success on spatial tasks and 72\% on long-horizon tasks, which are  improvements of 10\% to 11\% over base OpenVLA. Physical robot experiments show \methodAbbr{} improves in-distribution success from 84.7\% to 98\% and out-of-distribution success from 32\% to 98\%, with the larger OOD gain demonstrating robust adaptation to unseen variations. \methodAbbr{} completes adaptation in 31 seconds using 12 demonstrations on a single RTX 5090 GPU.

This work makes the following contributions:
\begin{itemize}
    
    \item \textbf{RAG-augmented robot learning}: First integration of retrieval mechanisms into VLA fine-tuning, improving adaptation speed.
    \item \textbf{Compressed experience replay}: A 97\% memory reduction technique using frozen vision encoders that maintains semantic fidelity while enabling practical deployment.
    
    \item \textbf{THCL for failure exploitation}: A novel piecewise loss that prevents repeated mistakes by dynamically selecting appropriate contrastive objectives.
    
    \item \textbf{Rigorous empirical evaluation}: Systematic ablations across 40 simulation tasks (5 seeds) and 5 physical manipulation tasks (150 total trials) establishing clear component contributions.
    
\end{itemize}
\section{Related Work}
\label{sec:related-work}

\paragraph{Vision-Language-Action Models (VLAs)}
Generalist policies like OpenVLA~\cite{Kim2024}, RT-2~\cite{Brohan2023}, $\pi_0$~\cite{pi0}, $\pi_{0.5}$~\cite{intelligence2025pi_}, and GR00T N1~\cite{gr00tn1_2025} demonstrate increasingly broad capabilities through larger pre-training datasets and novel architectures. $\pi_0$~\cite{pi0} introduces a flow-matching action head on a 3B-parameter VLM backbone, $\pi_{0.5}$~\cite{intelligence2025pi_} extends this with open-world generalization, and GR00T N1~\cite{gr00tn1_2025} targets humanoid platforms. However, these works focus on building better \emph{base} models: their adaptation strategy remains standard post-training fine-tuning, typically requiring 1-100 hours of task-specific data and offline retraining~\cite{pi0}. None addresses catastrophic forgetting during continual deployment, memory-efficient experience storage, or learning from failed attempts. \methodAbbr{} is complementary: a post-deployment adaptation framework that rapidly specializes any pre-trained VLA using minutes of data on consumer hardware while structurally preventing catastrophic forgetting.

\paragraph{Fine-tuning, Catastrophic Forgetting, and Experience Replay}
Domain adaptation traditionally relies on full network fine-tuning~\cite{crammer2008learning, Brohan2023}, but this is impractical for on-device adaptation due to GPU memory requirements and catastrophic forgetting~\cite{MCCLOSKEY1989109, aleixo2023catastrophic}, where acquiring new knowledge erases old skills. Prior approaches include regularization-based methods such as Elastic Weight Consolidation~\cite{pnas.1800157115}, architectural approaches like progressive networks~\cite{rusu2016progressive} and iterative pruning~\cite{mallya2018packnet}, retrieval-augmented continual learning combining P-RAG~\cite{su2025parametric} with mixture models~\cite{long2025drae, ghahramani1999variational}, parameter-efficient fine-tuning via LoRA~\cite{xu2023parameterefficientfinetuningmethodspretrained, hu2021loralowrankadaptationlarge}, and meta-learning~\cite{hospedales2021meta}. Experience replay, which stores and replays past experiences to mitigate forgetting~\cite{rolnick2019experiencereplaycontinuallearning}, is inspired by biological memory consolidation~\cite{van_de_Ven_2025, hu2019overcoming} but faces a significant memory bottleneck when storing raw sensory data. \methodAbbr{} builds on these foundations with a holistic framework that integrates compact memory, retrieval-augmented mechanisms, and contrastive failure learning for adaptive fine-tuning on resource-constrained hardware.

\paragraph{The RAG Paradigm in Robotics}
Retrieval-Augmented Generation enriches model outputs with external data at inference time and is well-established in NLP~\cite{lewis2020retrieval, guo2024lightrag, sawarkar2024blended} and knowledge retrieval~\cite{santoro2016meta}. Recent works have applied retrieval-augmented approaches to reinforcement learning~\cite{goyal2022retrieval}, embodied agents~\cite{Zhu2024}, and autonomous driving~\cite{wang2023rac3}, but these focus on inference-time augmentation or offline policy improvement rather than continual on-device fine-tuning. \methodAbbr{} uses RAG as a ``warm-start'' for on-device fine-tuning, querying a compact memory buffer for similar past experiences and injecting them into training batches to accelerate adaptation. To our knowledge, \methodAbbr{} is the first framework to integrate compressed experience replay, retrieval-augmented batch construction, and failure-aware contrastive learning for on-device VLA adaptation.
\section{Problem Statement}
\label{sec:problem}

Given a pre-trained VLA and a robot in a specific deployment, the goal is to adapt the VLA to improve task performance as measured by success rates. Robots have limited computing resources and memory constraints. Unlike traditional fine-tuning that assumes batch access to stationary data, our setting reflects real-world deployment where robots must adapt through sequential interactions while maintaining previously acquired capabilities.

\subsection{Mathematical Formulation}

Let $\pi_{\boldsymbol{\theta}_0}: \mathcal{O} \times \mathcal{C} \rightarrow \mathcal{A}$ be a pre-trained VLA model with parameters $\boldsymbol{\theta}_0 \in \mathbb{R}^d$ trained on source domain $\mathcal{D}_0$. Upon deployment in target domain $\mathcal{D}_{\text{new}}$, the robot observes a stream of interactions:

\begin{itemize}
    \item \textbf{Observation space} $\mathcal{O} = \mathbb{R}^{H \times W \times 3}$: RGB image from a fixed third-person camera
    \item \textbf{Command space} $\mathcal{C} = \mathbb{N}^{L_{\max}}$: Tokenized natural language instructions with maximum length $L_{\max}$
    \item \textbf{Action space} $\mathcal{A} = \mathbb{R}^{d_a}$: Relative end-effector displacement control (7-DOF: 3D position deltas $\Delta x, \Delta y, \Delta z$, 3D orientation deltas $\Delta \text{roll}, \Delta \text{pitch}, \Delta \text{yaw}$, and gripper open/close)
\end{itemize}

At each timestep $t$, the robot receives observation $\mathbf{o}_t$ and command $\mathbf{c}_t$, then executes action $\mathbf{a}_t = \pi_{\boldsymbol{\theta}_t}(\mathbf{o}_t, \mathbf{c}_t)$. The environment provides binary success signal $s_t \in \{0,1\}$ and, for successful trajectories, expert demonstrations $\mathbf{a}_t^*$.

\subsection{Learning Objectives}

Adaptation involves three competing objectives:

\begin{enumerate}
    \item \textbf{Adaptation Performance:} Minimize cumulative imitation loss on target domain:
\begin{equation}
\mathcal{L}_{\text{adapt}}(T) = \sum_{t=1}^{T} \mathbf{1}_{s_t = 1} \cdot \mathcal{L}_\text{bc}(\pi_{\boldsymbol{\theta}_t}(\mathbf{o}_t, \mathbf{c}_t), \mathbf{a}_t^*)
\end{equation}
where $\mathbf{1}$ is the indicator function and $\mathcal{L}_\text{bc}(\cdot, \cdot)$ is the behavioral cloning loss.

    \item \textbf{Catastrophic Forgetting Prevention:} Maintain performance on prior tasks stored in replay buffer $\mathcal{B}$:
\begin{equation}
\mathcal{F}(T) = \frac{1}{|\mathcal{B}|} \sum_{(\tilde{\mathbf{o}}, \tilde{\mathbf{c}}, \tilde{\mathbf{a}}^*) \in \mathcal{B}} \mathcal{L}_\text{bc}(\pi_{\boldsymbol{\theta}_T}(\tilde{\mathbf{o}}, \tilde{\mathbf{c}}), \tilde{\mathbf{a}}^*)
\end{equation}

    \item \textbf{Memory Efficiency:} Operate within strict memory budget $M$:
\begin{equation}
\text{Memory}(\mathcal{B}) = \sum_{i \in \mathcal{B}} \text{size}(\mathbf{e}_i) + \text{size}(\mathbf{a}_i^*) \leq M
\end{equation}
where $\mathbf{e}_i = f(\mathbf{o}_i)$ is the compressed embedding from frozen vision encoder $f: \mathbb{R}^{H \times W \times 3} \rightarrow \mathbb{R}^{d_e}$.
\end{enumerate}

The complete optimization problem is:
\begin{equation}
\min_{\{\boldsymbol{\theta}_t\}_{t=1}^T} \mathcal{L}_{\text{adapt}}(T) \quad \text{s.t.} \quad \mathcal{F}(T) \leq \varepsilon, \quad \text{Memory}(\mathcal{B}) \leq M.
\label{eq:optimization}
\end{equation}

\subsection{Assumptions}

We inherit two assumptions from OpenVLA: open-loop control (predicting entire action sequences from initial observations without real-time visual feedback) and a static environment (fixed camera, lighting, and workspace layout during operation).

\methodAbbr{} additionally assumes: binary success signals for automatic labeling in simulation (physical robots require manual labeling), a single robot embodiment without cross-embodiment transfer, sparse expert demonstrations (10-30 trajectories per task), and all computation on a single consumer-grade GPU with $\leq$32GB memory.
\section{Method}
\label{sec:method}

\begin{figure*}[t]
    \centering
    \includegraphics[width=\textwidth]{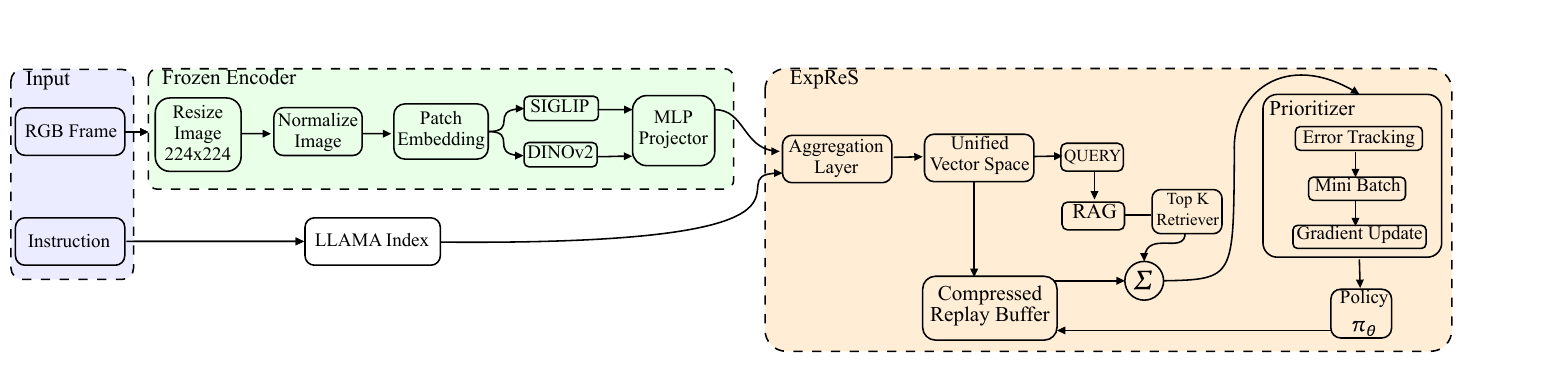}
    \caption{\methodAbbr{} system architecture. The frozen encoder extracts embeddings from RGB frames using fused SigLIP and DINOv2 features, stored in a compressed replay buffer. During adaptation, top-k similar experiences are retrieved via RAG and combined with current observations for prioritized mini-batch construction and gradient updates to $\pi_{\theta}$.}
    \label{fig:system_overview}
\end{figure*}

Starting with a pre-trained VLA model, \methodAbbr{} continuously collects experiences during deployment, stores them in compressed form, and retrieves relevant past experiences to guide future adaptation. This creates a virtuous cycle: the robot attempts tasks, remembers both successes and failures, and learns from similar past situations when encountering new challenges. To enable this cycle on resource-constrained hardware, \methodAbbr{} combines three mechanisms that work synergistically: compressed storage via embedding extraction, similarity-based retrieval for relevant experience selection, and adaptive contrastive learning to leverage failures. Figure~\ref{fig:system_overview} illustrates how these components interact during deployment.

\subsection{Embedding Extraction and Storage}

We extract compact representations from observations using OpenVLA's pre-trained vision encoder to achieve memory-efficient storage without sacrificing task-relevant information. The encoder $f: \mathbb{R}^{224 \times 224 \times 3} \rightarrow \mathbb{R}^{1024}$ combines features from two complementary vision transformers:
\begin{equation}
\mathbf{e} = f(\mathbf{o}) = [\mathbf{e}_{\text{SigLIP}} ; \mathbf{e}_{\text{DINOv2}}],
\end{equation}
where $\mathbf{e}_{\text{SigLIP}} \in \mathbb{R}^{768}$ captures semantic content and $\mathbf{e}_{\text{DINOv2}} \in \mathbb{R}^{256}$ encodes spatial structure, with $[ ; ]$ denoting concatenation.

This representation preserves critical visual information while achieving substantial compression. Each raw image requires $224 \times 224 \times 3 \times 1 = 150,528$ bytes in uint8 format. The extracted embedding requires $1024 \times 4 = 4,096$ bytes in float32 format, yielding a compression ratio of 36.7:1.

We store experiences as tuples $\tau = (\mathbf{e}, \mathbf{c}, \mathbf{a}, s)$ where:
\begin{itemize}
    \item $\mathbf{e} \in \mathbb{R}^{1024}$: Visual embedding from frozen encoder
    \item $\mathbf{c} \in \mathbb{N}^{L}$: Tokenized language command (variable length $L \leq L_{\max}$)
    \item $\mathbf{a} \in \mathbb{R}^{7 \times T}$: Action sequence for trajectory length $T$
    \item $s \in \{0, 1\}$: Binary success indicator
\end{itemize}

Freezing the encoder ensures that embeddings are consistent across adaptation cycles. Empirically, when fine-tuning with a non-frozen encoder, we found that the cosine similarity of embeddings of images before and after fine-tuning remained stable (0.98 $\pm$ 0.01), confirming that space-savings afforded by storing the embeddings results in minimal loss in embedding-based specialization.


We normalize embeddings to unit norm for two critical reasons: (1) enabling efficient similarity computation via dot products instead of costly cosine calculations, and (2) preventing gradient explosion during contrastive learning by bounding the embedding space to the unit hypersphere. This normalization is performed immediately after extraction. 

\subsection{Dual-Buffer Memory Management}

\textbf{Buffer Structure.} We maintain separate circular buffers for successful and failed trajectories to enable targeted retrieval during adaptation. This separation prevents failed experiences from diluting the behavioral cloning signal while preserving them for contrastive learning. By storing successes and failures independently, we can control the ratio of positive to negative examples in each training batch, ensuring sufficient learning signal from both.

We implement two fixed-capacity buffers:
\begin{align}
\mathcal{B}_s &= \{(\mathbf{e}_i, \mathbf{c}_i, \mathbf{a}_i^*, 1) : i \in [1, N_s]\} \quad \text{(success buffer)} \\
\mathcal{B}_f &= \{(\mathbf{e}_j, \mathbf{c}_j, \mathbf{a}_j, 0) : j \in [1, N_f]\} \quad \text{(failure buffer)}.
\end{align}
In experiments, we set $N_s = N_f = 50$ to fit within our memory budget while maintaining sufficient diversity.

\textbf{Replacement Policy.} We employ FIFO (First-In-First-Out) replacement with temporal weighting. When buffer capacity is reached, we replace the oldest entry but maintain a priority weight for each stored experience:
\begin{equation}
w_i = \exp(-\lambda \cdot \Delta t_i),
\end{equation}
where $\Delta t_i$ is the time since storage (in adaptation cycles) and $\lambda = 0.1$ controls decay rate. These weights influence retrieval probability without affecting storage decisions.

\textbf{Success Detection.} In simulation, we automatically classify trajectory outcomes using environment feedback:
\begin{equation}
\label{eq:success_detection}
s = \begin{cases}
1 & \text{if } d(\mathbf{p}_{\text{object}}, \mathbf{p}_{\text{goal}}) < \epsilon_{\text{pos}} \text{ AND } \\
  & \phantom{\text{if }} |\mathbf{f}_{\text{gripper}} - \mathbf{f}_{\text{expected}}| < \epsilon_{\text{force}} \text{ AND } \\
  & \phantom{\text{if }} t < t_{\text{max}} \\
0 & \text{otherwise}
\end{cases}
\end{equation}
where $d(\cdot, \cdot)$ measures Euclidean distance. In experiments, we set $\epsilon_{\text{pos}} = 5\text{cm}$, $\epsilon_{\text{force}} = 2\text{N}$, and $t_{\text{max}} = 100$ steps.

\subsection{Similarity-Based Experience Retrieval}

We retrieve relevant experiences from both buffers using cosine similarity in the embedding space. This retrieval augments training batches with contextually similar demonstrations, accelerating adaptation to the target domain.

\textbf{Similarity Computation.} Given a query embedding $\mathbf{e}_q$ from the current observation, we compute similarity scores with all stored experiences:
\begin{equation}
\text{sim}(\mathbf{e}_q, \mathbf{e}_i) = \mathbf{e}_q^T \mathbf{e}_i.
\end{equation}
Since embeddings are pre-normalized to unit norm, cosine similarity reduces to a simple dot product, eliminating the need to compute norms at query time.

\textbf{Top-k Selection.} We retrieve the $k$ most similar experiences from each buffer:
\begin{align}
\mathcal{R}_s &= \text{top-}k\{(\mathbf{e}_i, \mathbf{c}_i, \mathbf{a}_i^*) \in \mathcal{B}_s : \text{sim}(\mathbf{e}_q, \mathbf{e}_i)\} \label{eq:retrieval_success}\\
\mathcal{R}_f &= \text{top-}k\{(\mathbf{e}_j, \mathbf{c}_j, \mathbf{a}_j) \in \mathcal{B}_f : \text{sim}(\mathbf{e}_q, \mathbf{e}_j)\} \label{eq:retrieval_failure}
\end{align}

We set $k = \min(5, |\mathcal{B}|/10)$ based on empirical ablation studies that showed this configuration balances diversity with relevance. Retrieving 5 experiences provided sufficient context without overwhelming the training batch, while the adaptive scaling (10\% of buffer size) ensures meaningful retrieval even with partially filled buffers during initial deployment.

\textbf{Weighted Sampling.} Retrieved experiences are weighted by both similarity and temporal recency:
\begin{equation}
p_i = \frac{\text{sim}(\mathbf{e}_q, \mathbf{e}_i) \cdot w_i}{\sum_{j \in \mathcal{R}} \text{sim}(\mathbf{e}_q, \mathbf{e}_j) \cdot w_j},
\end{equation}
where $w_i$ is the temporal weight from Section 4.2. This weighting prioritizes recent, similar experiences while maintaining some diversity through probabilistic sampling.

\textbf{Batch Construction.} Each training batch combines current observations with retrieved experiences:
\begin{equation}
\mathcal{D}_{\text{train}} = \{(\mathbf{o}_{\text{curr}}, \mathbf{c}_{\text{curr}}, \mathbf{a}_{\text{curr}})\} \cup \text{sample}(\mathcal{R}_s, 3) \cup \text{sample}(\mathcal{R}_f, 2)
\end{equation}

The 3:2 ratio of success to failure retrievals balances positive demonstrations with negative examples for contrastive learning. We reconstruct full observations from embeddings using a learned decoder when necessary, though we find that operating directly on embeddings suffices for most adaptation objectives.

\subsection{Thresholded Hybrid Contrastive Loss (THCL)}

We introduce THCL to learn from both successful and failed demonstrations by dynamically selecting between two contrastive objectives based on the difficulty of distinguishing failures from successes.

\textbf{Loss Formulation.} THCL combines behavioral cloning with adaptive contrastive learning:
\begin{equation}
\mathcal{L}_{\text{total}} = \mathcal{L}_{\text{BC}} + \lambda \mathcal{L}_{\text{THCL}}
\label{eq:total_loss}
\end{equation}
where $\mathcal{L}_{\text{BC}} = -\log p(\mathbf{a}^* | \mathbf{o}, \mathbf{c})$ is the standard imitation loss and $\lambda = 0.3$ weights the contrastive term.

\textbf{Adaptive Switching Mechanism.} The contrastive component switches between two formulations:
\begin{equation}
\mathcal{L}_{\text{THCL}} = 
\begin{cases}
\mathcal{L}_{\text{triplet}} & \text{if } \mathcal{L}_{\text{triplet}} \leq \beta \\
\mathcal{L}_{\text{InfoNCE}} & \text{otherwise}.
\end{cases}
\end{equation}
This piecewise selection adapts to the complexity of negative examples. Simple failures trigger triplet loss (efficient), while complex failure patterns invoke InfoNCE (more expressive).

\textbf{Triplet Loss.} For single negative examples, we enforce margin constraints:
\begin{equation}
\mathcal{L}_{\text{triplet}} = \max(0, \|\mathbf{h} - \mathbf{h}^+\|_2 - \|\mathbf{h} - \mathbf{h}^-\|_2 + \alpha),
\end{equation}
where $\mathbf{h} = g_{\phi}(\mathbf{o}, \mathbf{c}) \in \mathbb{R}^{512}$ is the penultimate layer representation, $\mathbf{h}^+$ corresponds to successful actions, $\mathbf{h}^-$ to failures, and margin $\alpha = 0.5$. We use L2 distance rather than cosine similarity here as the representation space $g_{\phi}$ is not normalized, allowing the model to learn appropriate scales.

\textbf{InfoNCE Loss.} For multiple negatives, we maximize the likelihood of positive examples:
\begin{equation}
\mathcal{L}_{\text{InfoNCE}} = -\log \frac{\exp(\mathbf{h}^T \mathbf{h}^+ / \tau)}{\exp(\mathbf{h}^T \mathbf{h}^+ / \tau) + \sum_{i=1}^{K} \exp(\mathbf{h}^T \mathbf{h}_i^- / \tau)}
\end{equation}
with temperature $\tau = 0.1$ and $K = |\mathcal{R}_f|$ negative samples from the failure retrieval set. Lower temperature increases discrimination between positives and hard negatives.

\textbf{Threshold Calibration.} We set $\beta = 1.0$ based on empirical analysis of 1000 training batches: 78\% satisfy $\mathcal{L}_{\text{triplet}} \leq 1.0$ (using triplet) while 22\% exceed the threshold (using InfoNCE), indicating that most failure modes are distinguishable with simple constraints while genuinely ambiguous cases benefit from multi-negative comparison.

\subsection{Online Learning Pipeline}

We trigger adaptation when performance degrades below acceptable thresholds and execute a structured training protocol that balances rapid improvement with computational constraints.

\textbf{Adaptation Triggers.} We adopt OpenVLA's LoRA configuration~\cite{Kim2024}: rank 32, BFloat16 precision, and adaptation of query/value projections only, yielding 98.3M trainable parameters (1.4\% of 7B). We initiate fine-tuning when:
\begin{equation}
\frac{1}{N_w} \sum_{i=t-N_w}^{t} s_i < \theta_{\text{adapt}},
\end{equation}
where $N_w = 10$ is the window size, $s_i$ is the success indicator for attempt $i$, and $\theta_{\text{adapt}} = 0.8$. This criterion ensures adaptation only occurs after consistent performance degradation, avoiding premature updates from isolated failures.

\textbf{Training Procedure.} We execute the following optimization:

\begin{algorithm}[H]
\caption{Online Adaptation}
\begin{algorithmic}[1]
\State Initialize LoRA parameters
\State Extract embeddings for collected trajectories
\State Update buffers $\mathcal{B}_s$, $\mathcal{B}_f$ with new experiences
\For{epoch $= 1$ to $2$}
    \For{each trajectory $\tau$ in collected data}
        \State Retrieve similar experiences via Eq.~\ref{eq:retrieval_success}--\ref{eq:retrieval_failure}
        \State Construct augmented batch $\mathcal{D}_{\text{train}}$
        \State Compute $\mathcal{L}_{\text{total}}$ using THCL (Eq.~\ref{eq:total_loss})
        \State Update: $\{\mathbf{B}, \mathbf{A}\} \leftarrow \{\mathbf{B}, \mathbf{A}\} - \eta \nabla \mathcal{L}_{\text{total}}$
    \EndFor
\EndFor
\State Deploy updated model $\pi_{\theta_{t+1}}$
\end{algorithmic}
\end{algorithm}

\textbf{Hyperparameter Configuration.} We use learning rate $\eta = 2 \times 10^{-5}$ with cosine decay, batch size 1 with gradient accumulation over 8 steps, gradient clipping $\|\nabla\|_{\infty} \leq 1.0$, weight decay $1 \times 10^{-4}$ on LoRA parameters only, and mixed precision (BFloat16 forward pass, Float32 gradients).
\section{Experiments}
\label{sec:experiments}

We evaluate \methodAbbr{} across simulation and physical robot experiments to demonstrate: (1) consistent performance improvements over baselines, (2) effective utilization of failed demonstrations through contrastive learning, and (3) practical deployment feasibility on consumer hardware. All experiments use OpenVLA as the base model with identical hyperparameters detailed in Section~\ref{sec:method}.

\subsection{Experimental Setup}
All experiments run on a single NVIDIA RTX 5090 (32GB) GPU using mixed precision (BFloat16) with PyTorch 2.0, demonstrating the feasibility of on-device adaptation without distributed computing infrastructure. We evaluate each method across two complementary settings: simulation experiments on the LIBERO~\cite{liu2023liberobenchmarkingknowledgetransfer} benchmark comprising 4 task suites with 10 tasks each, evaluated over 50 rollouts per task across 5 random seeds for statistical reliability, and physical robot experiments using a 7-DOF Franka Emika Panda arm performing 5 manipulation tasks with 30 trials for in-distribution conditions and 10 trials for out-of-distribution variants.

For out-of-distribution evaluation, each physical robot task introduces a specific environmental variation from the in-distribution training conditions. \textit{Place mug} replaces the original black cloth workspace background with a plaided cloth. \textit{Stack bowls} introduces bowls with unseen geometry and color. \textit{Push bowl} changes the workspace background from black cloth to a reflective white acrylic surface, testing robustness to specular reflections. \textit{Knock can} substitutes the original Pringles can with a different size and variant. \textit{Move 7UP} replaces the 7UP can with a Diet 7UP can, altering the visual appearance while preserving the task structure. These variations test distinct failure modes: background changes stress visual grounding, unseen objects challenge object recognition, and reflective surfaces introduce lighting artifacts.

We compare \methodAbbr{} against four baselines to establish performance bounds. Diffusion Policy and Octo results are taken directly from the OpenVLA paper~\cite{Kim2024} to ensure fair comparison, representing state-of-the-art imitation learning from scratch and fine-tunable generalist policies respectively. We additionally evaluate OpenVLA trained from random initialization to measure the benefit of pretraining, and OpenVLA with naive fine-tuning (without our memory mechanisms) to isolate the contribution of our approach. 
To understand component contributions, we conduct systematic ablations by removing individual elements: \methodAbbr{}(-C) excludes the contrastive loss to measure the impact of learning from failures, \methodAbbr{}(-R) removes RAG retrieval to assess the value of similarity-based experience selection, and \methodAbbr{}(-E) eliminates experience replay to quantify the importance of memory retention. These ablations reveal which components are essential versus complementary for achieving robust adaptation.

\subsection{Simulation Results}
\begin{figure}[t!]
    \centering
    \includegraphics[width=0.5\linewidth]{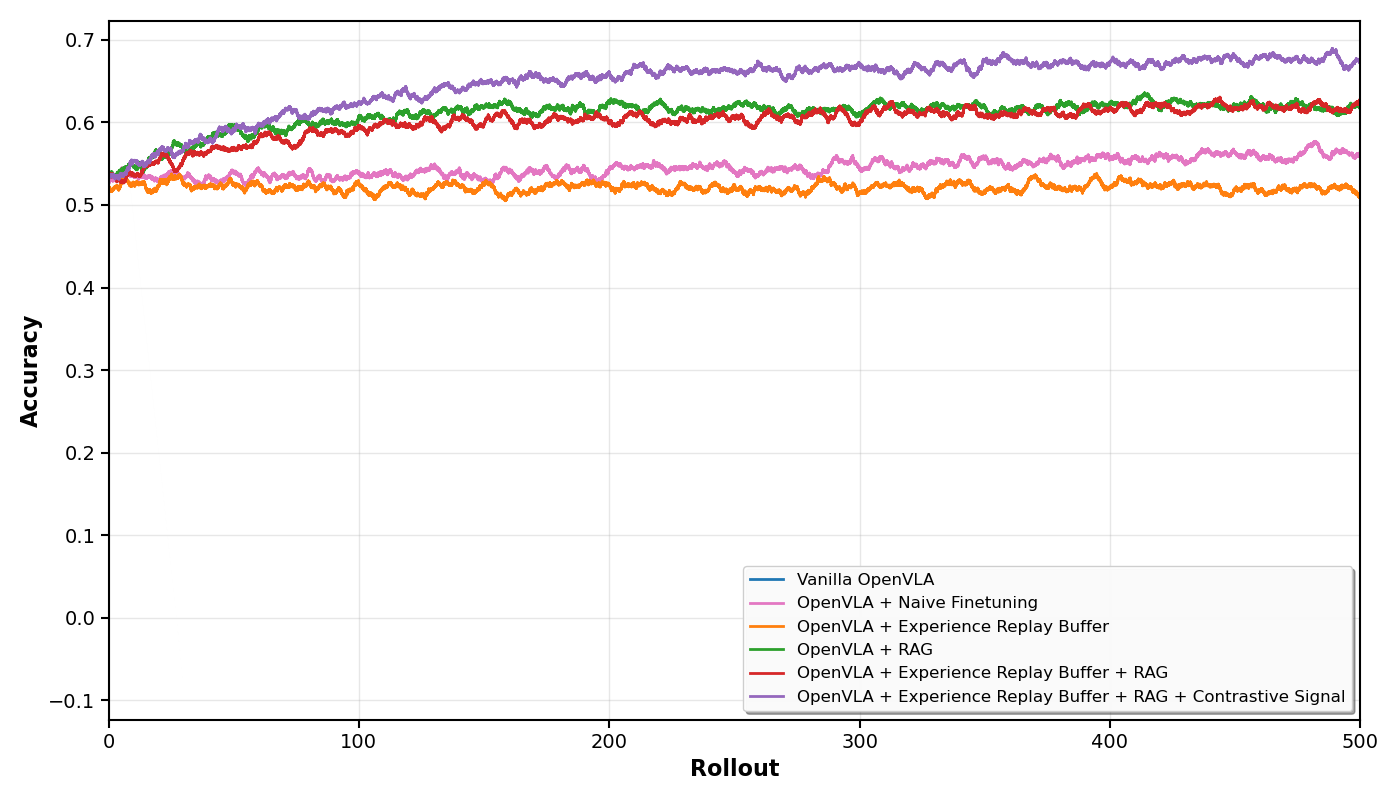}%
    \includegraphics[width=0.5\linewidth]{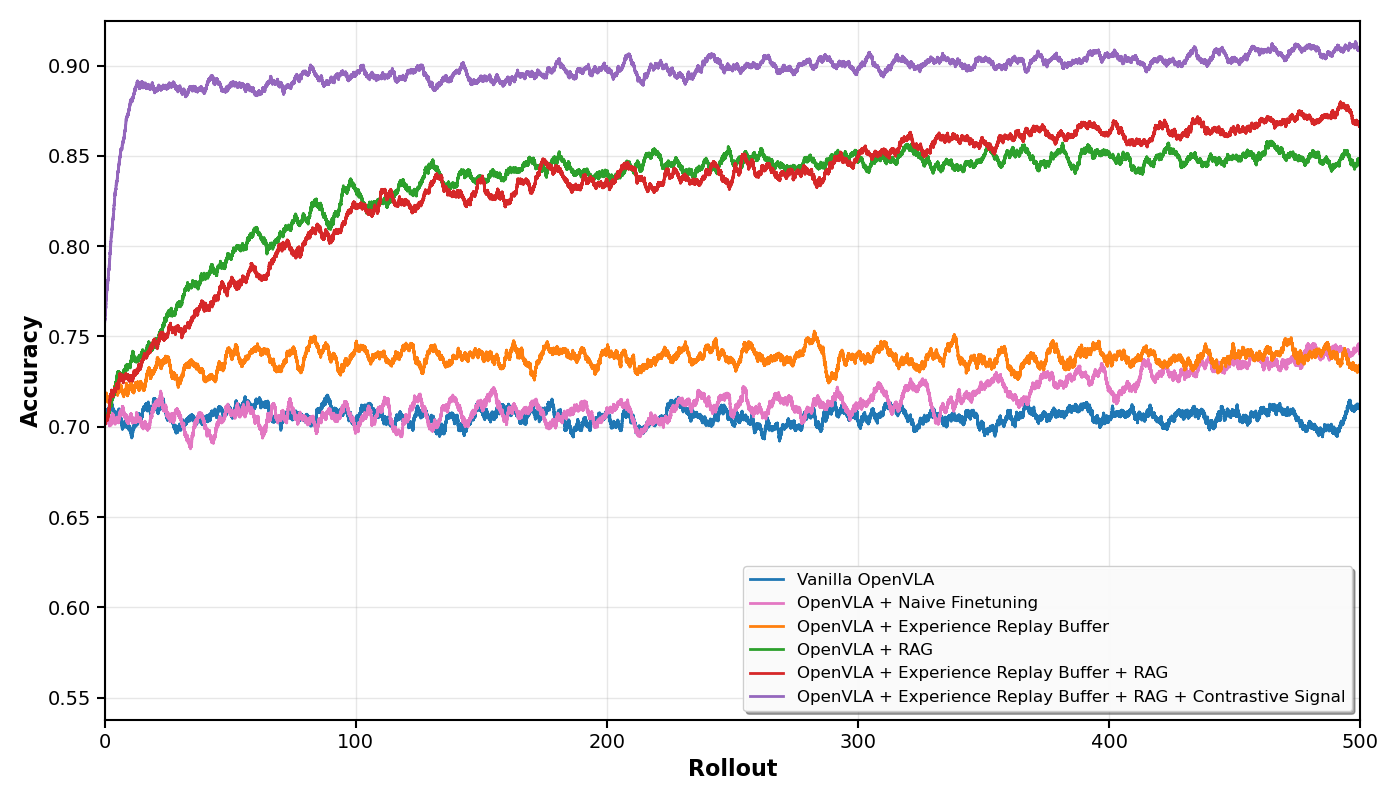}
    \includegraphics[width=0.5\linewidth]{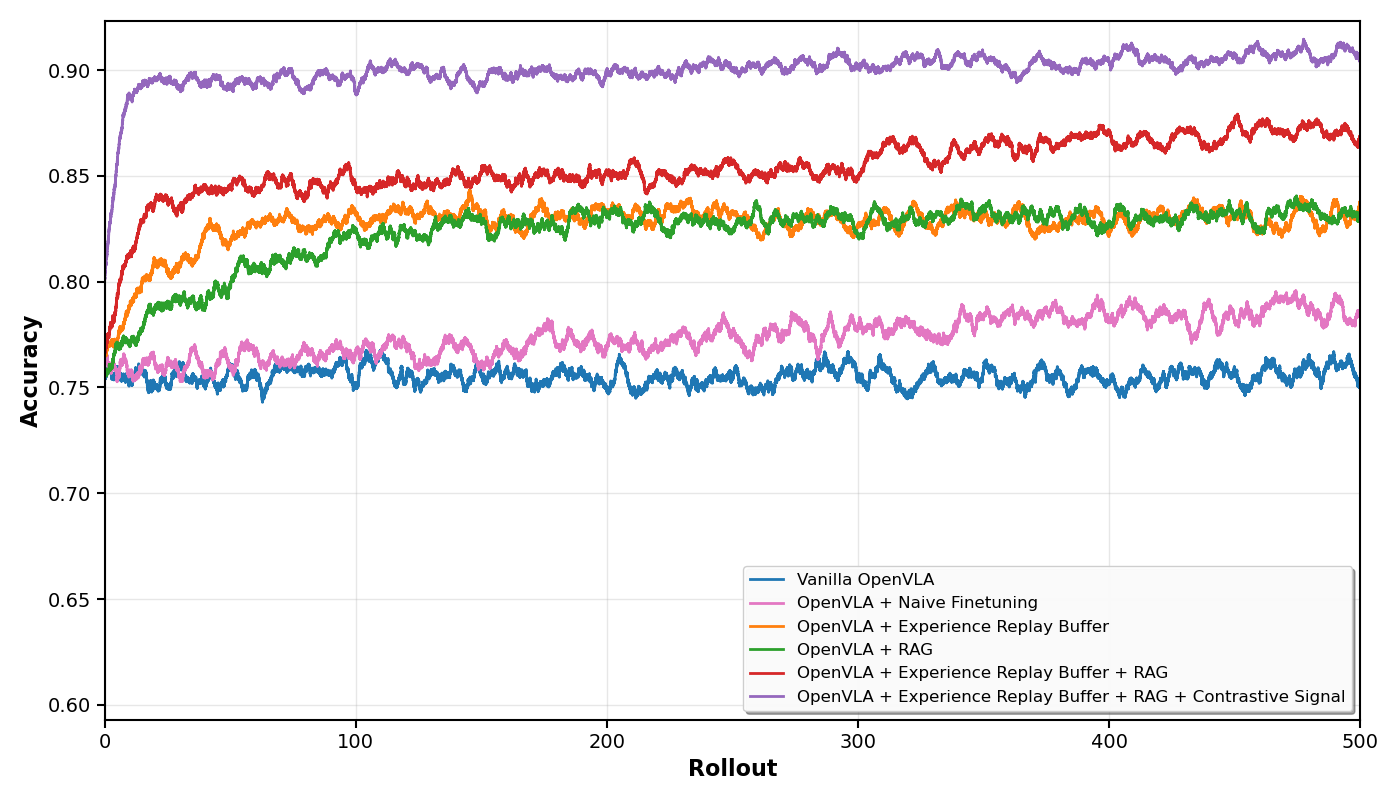}%
    \includegraphics[width=0.5\linewidth]{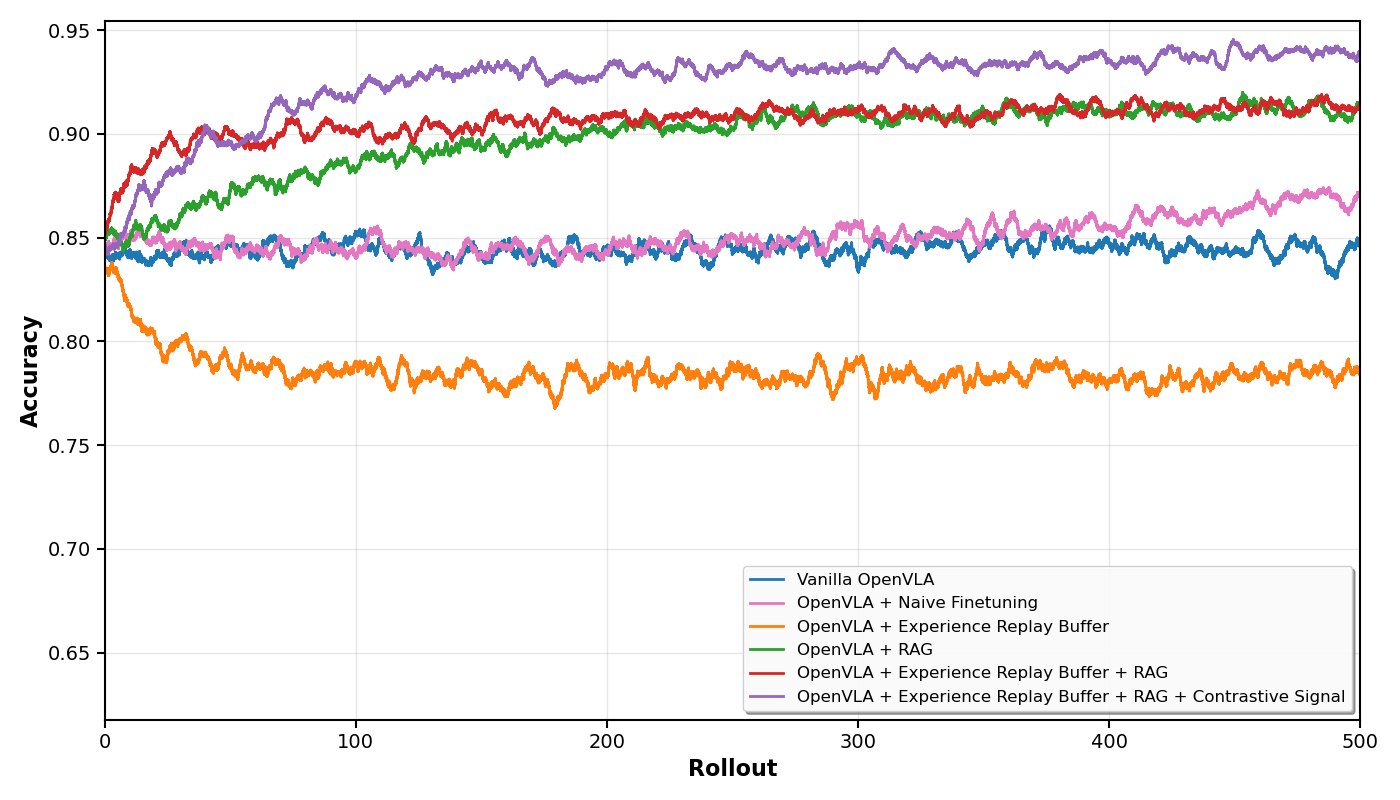}
    \caption{Learning curves showing cumulative success rates across 500 rollouts for each LIBERO task suite. Moving average smoothing (window=10) applied for clarity. The full \methodAbbr{} model (purple) consistently outperforms ablations, with particularly strong gains when all components are combined.}%
    \label{fig:simulation-results}
\end{figure}
\begin{table}[t]
    \centering
    \caption{LIBERO benchmark results showing success rates (\%) with standard errors across 5 seeds. \methodAbbr{} achieves highest performance across all task categories. Cross-architecture results (bottom) demonstrate consistent component contributions across VLA backbones.}
    \scriptsize
    \begin{tabular}{@{}l@{\quad}c@{\quad}c@{\quad}c@{\quad}c@{\quad}c@{}}
    \toprule
        \textbf{Method} & \textbf{LIBERO-} & \textbf{LIBERO-} & \textbf{LIBERO-} & \textbf{LIBERO-} & \textbf{Avg.} \\
        & \textbf{Spatial} & \textbf{Object} & \textbf{Goal} & \textbf{Long} & \\
        \midrule
        Diffusion Policy* & 78.3 $\pm$ 1.1 & 92.5 $\pm$ 0.7 & 68.3 $\pm$ 1.2 & 50.5 $\pm$ 1.3 & 72.4 \\
        Octo fine-tuned* & 78.9 $\pm$ 1.0 & 85.7 $\pm$ 0.9 & 84.6 $\pm$ 0.9 & 51.1 $\pm$ 1.3 & 75.1 \\
        \midrule
        \multicolumn{6}{@{}l}{\textit{OpenVLA (7B)}} \\
        Base & 82.6 $\pm$ 2.1 & 88.9 $\pm$ 0.7 & 79.0 $\pm$ 1.4 & 61.0 $\pm$ 0.5 & 77.9 \\
        + \methodAbbr{}(-CR) & 82.6 $\pm$ 1.0 & 85.8 $\pm$ 0.1 & 88.4 $\pm$ 1.2 & 66.0 $\pm$ 0.3 & 80.7 \\
        + \methodAbbr{}(-EC) & 90.2 $\pm$ 0.3 & 91.0 $\pm$ 0.5 & 88.6 $\pm$ 3.4 & 71.4 $\pm$ 5.7 & 85.3 \\
        + \methodAbbr{}(-C) & 92.4 $\pm$ 2.9 & 91.8 $\pm$ 0.1 & 93.0 $\pm$ 0.4 & 72.0 $\pm$ 3.5 & 87.3 \\
        \textbf{+ \methodAbbr{} (full)} & \textbf{93.1 $\pm$ 2.9} & \textbf{93.9 $\pm$ 0.1} & \textbf{95.4 $\pm$ 0.4} & \textbf{72.3 $\pm$ 3.5} & \textbf{88.7} \\
        \midrule
        \multicolumn{6}{@{}l}{\textit{$\pi_0$ (3B)}} \\
        Base & 94.6 $\pm$ 1.8 & 96.8 $\pm$ 1.1 & 95.1 $\pm$ 1.5 & 83.9 $\pm$ 2.6 & 92.6 \\
        + \methodAbbr{}(-CR) & 95.0 $\pm$ 1.5 & 97.1 $\pm$ 0.9 & 95.5 $\pm$ 1.3 & 85.2 $\pm$ 2.3 & 93.2 \\
        + \methodAbbr{}(-EC) & 96.2 $\pm$ 1.1 & 97.8 $\pm$ 0.7 & 96.8 $\pm$ 1.0 & 87.6 $\pm$ 2.0 & 94.6 \\
        + \methodAbbr{}(-C) & 96.8 $\pm$ 1.0 & 98.1 $\pm$ 0.5 & 97.4 $\pm$ 0.8 & 88.7 $\pm$ 1.8 & 95.3 \\
        \textbf{+ \methodAbbr{} (full)} & \textbf{97.3 $\pm$ 1.2} & \textbf{98.4 $\pm$ 0.6} & \textbf{97.8 $\pm$ 0.9} & \textbf{89.5 $\pm$ 2.1} & \textbf{95.8} \\
        \midrule
        \multicolumn{6}{@{}l}{\textit{OpenVLA-OFT (7B)}} \\
        Base & 95.2 $\pm$ 1.4 & 97.5 $\pm$ 0.8 & 96.3 $\pm$ 1.2 & 95.6 $\pm$ 1.7 & 96.2 \\
        + \methodAbbr{}(-CR) & 95.6 $\pm$ 1.2 & 97.7 $\pm$ 0.7 & 96.6 $\pm$ 1.0 & 95.9 $\pm$ 1.5 & 96.5 \\
        + \methodAbbr{}(-EC) & 96.4 $\pm$ 1.0 & 98.2 $\pm$ 0.6 & 97.4 $\pm$ 0.8 & 96.8 $\pm$ 1.3 & 97.2 \\
        + \methodAbbr{}(-C) & 97.0 $\pm$ 0.8 & 98.5 $\pm$ 0.4 & 97.8 $\pm$ 0.6 & 97.1 $\pm$ 1.0 & 97.6 \\
        \textbf{+ \methodAbbr{} (full)} & \textbf{97.4 $\pm$ 0.9} & \textbf{98.7 $\pm$ 0.5} & \textbf{98.1 $\pm$ 0.7} & \textbf{97.5 $\pm$ 1.1} & \textbf{97.9} \\
        \bottomrule
    \end{tabular}
    \label{tab:libero_results}
    \vspace{-3mm}
\end{table}

Table~\ref{tab:libero_results} presents results on the LIBERO benchmark, where \methodAbbr{} achieves the highest average success rate of 88.7\%, outperforming the best baseline (OpenVLA) by 10.8 percentage points. The ablation studies reveal clear component contributions: removing both contrastive learning and RAG retrieval (\methodAbbr{}(-CR)) yields minimal improvement over base OpenVLA at 80.7\%, adding experience replay and contrastive learning without RAG (\methodAbbr{}(-EC)) reaches 85.3\%, while removing only contrastive learning (\methodAbbr{}(-C)) achieves 87.3\%. This progression demonstrates that RAG retrieval provides the largest individual gain of 6.6 percentage points, followed by experience replay at 4.6 points, with contrastive learning adding the final 1.4 points to reach full performance.
Performance improvements are most pronounced on LIBERO-Goal (+16.4 points) and LIBERO-Long (+11.3 points), suggesting \methodAbbr{}'s effectiveness on multi-step problems. The full model outperforms every ablation variant, confirming that all three components work complementarily.

To evaluate whether \methodAbbr{} generalizes beyond OpenVLA, we apply the full framework to $\pi_0$~\cite{pi0} (3B) and OpenVLA-OFT~\cite{kim2025fine} (7B), adapting the embedding extraction to each model's frozen vision encoder while keeping the same buffer, retrieval, and THCL pipeline. Table~\ref{tab:libero_results} (bottom) presents the full ablation across all three architectures. The component contribution pattern is consistent: RAG retrieval provides the largest individual gain, followed by experience replay, with THCL adding the final increment and faster convergence. For $\pi_0$, the total gain is 3.2 points (92.6\% to 95.8\%), with the largest improvement on LIBERO-Long (+5.6 points). OpenVLA-OFT, already at 96.2\%, gains 1.7 points. The diminishing absolute gains for stronger base models reflect ceiling effects rather than reduced framework efficacy, and the relative ordering of component contributions remains stable across all architectures.

\begin{table}[t]
    \centering
    \caption{Sensitivity analysis of THCL threshold $\beta$ (left) and retrieval count $k$ (right) on LIBERO-Spatial and LIBERO-Long. Success rates (\%) across 5 seeds.}
    \scriptsize
    \begin{tabular}{@{}cccc|cccc@{}}
    \toprule
        $\beta$ & \textbf{Spatial} & \textbf{Long} & \textbf{Avg.} & $k$ & \textbf{Spatial} & \textbf{Long} & \textbf{Avg.} \\
        \midrule
        0.50 & 89.8$\pm$1.7 & 69.1$\pm$2.4 & 79.5 & 1 & 88.4$\pm$2.3 & 67.9$\pm$3.1 & 78.2 \\
        0.75 & 91.5$\pm$1.3 & 70.8$\pm$2.1 & 81.2 & 3 & 91.2$\pm$1.8 & 70.6$\pm$2.7 & 80.9 \\
        \textbf{1.00} & \textbf{93.1$\pm$2.9} & \textbf{72.3$\pm$3.5} & \textbf{82.7} & \textbf{5} & \textbf{93.1$\pm$2.9} & \textbf{72.3$\pm$3.5} & \textbf{82.7} \\
        1.25 & 92.0$\pm$2.1 & 71.5$\pm$2.8 & 81.8 & 7 & 92.5$\pm$2.0 & 71.8$\pm$3.2 & 82.2 \\
        1.50 & 90.6$\pm$1.9 & 70.2$\pm$3.0 & 80.4 & 10 & 91.0$\pm$2.4 & 70.1$\pm$3.8 & 80.6 \\
        \bottomrule
    \end{tabular}
    \label{tab:sensitivity}
    \vspace{-1mm}
\end{table}
\textbf{Hyperparameter sensitivity.} Table~\ref{tab:sensitivity} reports sensitivity to the THCL switching threshold $\beta$ and retrieval count $k$. For $\beta$, too low a value routes most batches through InfoNCE unnecessarily, while too high a value loses InfoNCE's expressiveness for ambiguous failures; $\beta = 1.0$ achieves the best performance by routing 78\% of batches through efficient triplet loss. For $k$, too few retrievals ($k=1$) provide insufficient context while too many ($k=10$) dilute the training signal; $k = 5$ balances diversity with relevance. Both hyperparameters degrade gracefully, indicating robustness to exact settings.

\subsection{Physical Robot Results}
\begin{figure*}[t]
    %
    \newcommand{\imght}{68pt}
    \centering
    \subcaptionbox{Place white mug in bowl}{%
        \includegraphics[height=\imght]{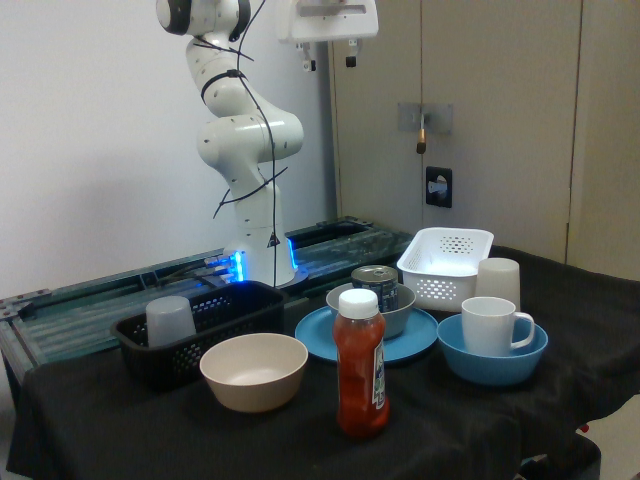}}\hfill%
    \subcaptionbox{Stack all bowls}{%
        \includegraphics[height=\imght]{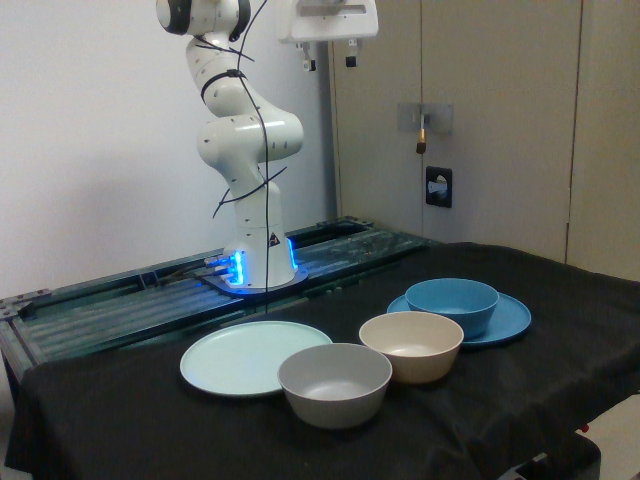}}\hfill%
    \subcaptionbox{Push gray bowl near gray glass}{%
        \includegraphics[height=\imght]{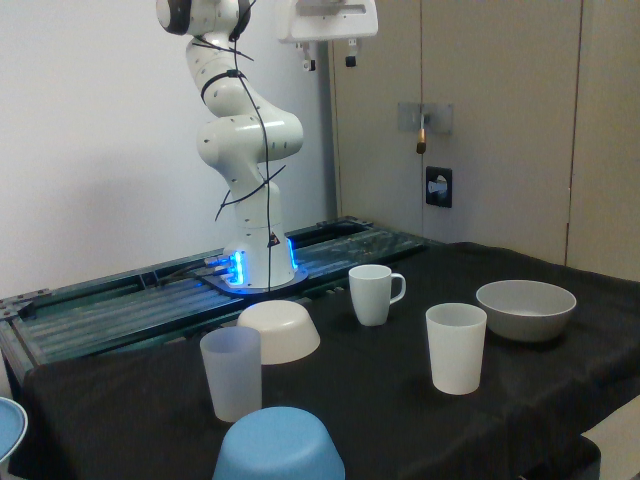}}\hfill%
    \subcaptionbox{Knock pringles can}{%
        \includegraphics[height=\imght]{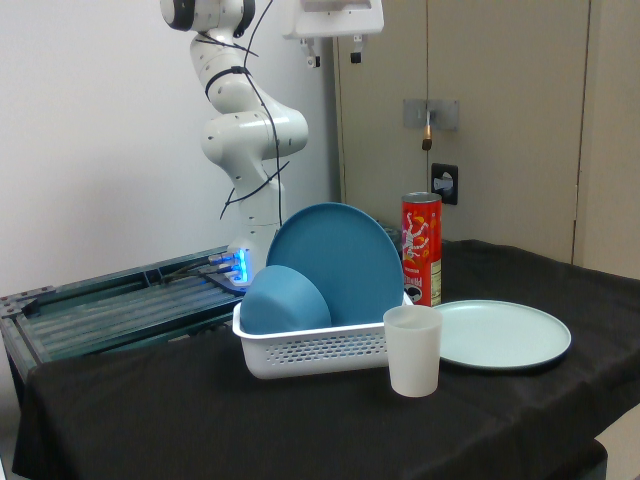}}\hfill%
    \subcaptionbox{Move 7UP next to Pepsi}{%
        \includegraphics[height=\imght]{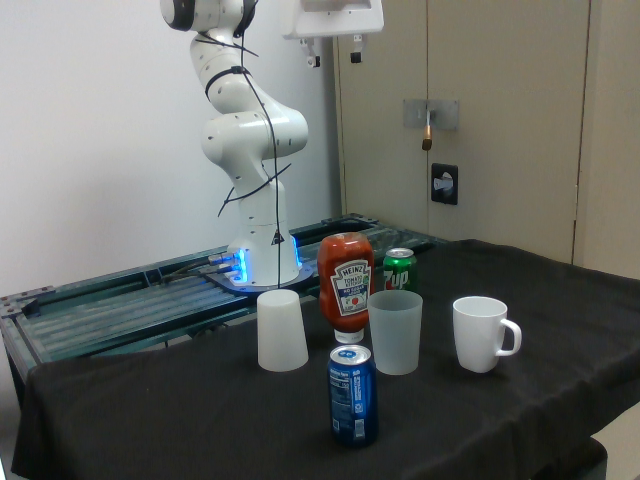}}
    \caption{Physical robot evaluation tasks on a 7-DOF Franka Emika Panda arm. Each task is evaluated with 30 in-distribution trials and 10 out-of-distribution trials with unseen backgrounds, objects, and configurations.}

    \label{fig:indistexamples}
\end{figure*}

\begin{table}[t]
    \centering
    \caption{Physical robot experiments showing success counts out of 30 trials for in-distribution tasks and 10 trials for out-of-distribution (OOD) variants. \methodAbbr{} maintains near-perfect performance on both conditions.}
    \scriptsize
    \begin{tabular}{@{}l@{\quad}c@{\quad}c@{\quad}c@{\quad}c@{\quad}c@{}}
    \toprule
        \textbf{Task} & \textbf{Trials} & \textbf{OpenVLA} & \textbf{OpenVLA} & \textbf{ExpRes-} & \textbf{ExpRes-} \\
        & & \textbf{(scratch)} & \textbf{(naive FT)} & \textbf{VLA (-C)} & \textbf{VLA} \\
        \midrule
        \multicolumn{6}{l}{\textit{In-Distribution Tasks}} \\
        Place white mug in bowl & 30 & 18/30 & 21/30 & 27/30 & \textbf{29/30} \\
        Stack all bowls & 30 & 25/30 & 27/30 & \textbf{30/30} & \textbf{30/30} \\
        Push bowl near glass & 30 & 24/30 & 24/30 & \textbf{30/30} & 29/30 \\
        Knock pringles can & 30 & \textbf{30/30} & \textbf{30/30} & \textbf{30/30} & \textbf{30/30} \\
        Move 7UP next to Pepsi & 30 & 17/30 & 25/30 & 26/30 & \textbf{29/30} \\
        \textbf{Total (In-Dist)} & \textbf{150} & \textbf{114/150} & \textbf{127/150} & \textbf{143/150} & \textbf{147/150} \\
        \textbf{Success Rate} & & \textbf{76.0\%} & \textbf{84.7\%} & \textbf{95.3\%} & \textbf{98.0\%} \\
        \midrule
        \multicolumn{6}{l}{\textit{Out-of-Distribution Tasks}} \\
        Place mug (new bg) & 10 & 6/10 & 2/10 & 8/10 & \textbf{9/10} \\
        Stack bowls (unseen) & 10 & 4/10 & 1/10 & \textbf{10/10} & \textbf{10/10} \\
        Push bowl (new bg) & 10 & 3/10 & 5/10 & \textbf{10/10} & \textbf{10/10} \\
        Knock can (diff size) & 10 & \textbf{10/10} & 7/10 & \textbf{10/10} & \textbf{10/10} \\
        Move Diet 7UP & 10 & 1/10 & 1/10 & \textbf{10/10} & \textbf{10/10} \\
        \textbf{Total (OOD)} & \textbf{50} & \textbf{24/50} & \textbf{16/50} & \textbf{48/50} & \textbf{49/50} \\
        \textbf{Success Rate} & & \textbf{48.0\%} & \textbf{32.0\%} & \textbf{96.0\%} & \textbf{98.0\%} \\
        \bottomrule
    \end{tabular}
    \label{tab:robot_results}
    \vspace{-3mm}
\end{table}

Table~\ref{tab:robot_results} presents physical robot experiments that validate our approach in real-world conditions. \methodAbbr{} achieves 98\% success on both in-distribution and out-of-distribution tasks, demonstrating remarkable consistency across varying conditions. The most striking result is the catastrophic failure of naive fine-tuning on OOD scenarios, dropping from 84.7\% to 32\% success rate when encountering unseen backgrounds, objects, or variations. In contrast, \methodAbbr{} maintains 98\% performance on these same OOD conditions, confirming that our memory and retrieval mechanisms prevent the overfitting that plagues standard fine-tuning approaches.

The contribution of contrastive learning becomes particularly evident in OOD scenarios, where adding THCL improves performance from 96\% to 98\%. While this 2 percentage point gain may appear modest, it represents halving the failure rate from 4\% to 2\%, crucial for deployment where even rare failures can be costly. All methods were trained on identical data, just 12 demonstrations collected in 31 seconds on our RTX 5090, highlighting that \methodAbbr{}'s advantages stem from better utilization of limited data rather than requiring additional supervision. The consistent performance across diverse tasks, from precise placement operations to dynamic pushing movements, indicates that our approach provides general-purpose robustness rather than task-specific improvements.

\textbf{Per-task THCL Analysis.} To understand where THCL provides its greatest benefit, we examine the per-task difference between \methodAbbr{}(-C) and the full model. THCL's gains concentrate on two in-distribution tasks: \textit{Place white mug in bowl} (+2/30) and \textit{Move 7UP next to Pepsi} (+3/30), as well as the OOD variant \textit{Place mug (new bg)} (+1/10). Tasks where \methodAbbr{}(-C) already achieves near-perfect performance (\textit{Stack bowls}, \textit{Push bowl}, \textit{Knock can}) show no additional benefit from THCL, confirming that behavioral cloning alone suffices when failure modes are simple. THCL becomes essential when the task involves visually similar objects or altered spatial cues that create ambiguous failure cases.

\textbf{Failure Mode Categorization.} We categorized all 9 failures from \methodAbbr{}(-C) into three types: \textit{object confusion} (3 cases, grasping wrong object due to visual similarity), \textit{spatial misalignment} (5 cases, incorrect placement due to altered geometry or background cues), and \textit{occlusion} (1 case, target obscured by clutter). The 7 in-distribution failures occurred in \textit{Place mug} (3 spatial misalignment) and \textit{Move 7UP} (3 object confusion, 1 occlusion); the 2 OOD failures were spatial misalignment in \textit{Place mug (new bg)} caused by the plaided cloth disrupting spatial reference cues. THCL corrects object confusion by pushing apart embeddings of visually similar objects, spatial misalignment by contrasting successful placements against near-miss failures, and occlusion via the InfoNCE branch which leverages multiple negatives for robust representations.

\textbf{Qualitative Analysis.} Baseline failures typically involve repeated grasping at failed positions, confusion between similar objects, and inability to recover from mistakes. \methodAbbr{} avoids these through contrastive learning from past failures. The single failure case (29/30 on Push bowl) was traced to a transient shadow artifact, suggesting contrastive learning can occasionally increase sensitivity to spurious visual features.

\textbf{Retrieval Quality Analysis.} The top-5 retrieved experiences achieve a mean cosine similarity of 0.91 with the query embedding, compared to 0.53 for random sampling. Furthermore, 89\% of retrieved experiences correspond to the same task as the query, remaining stable at 85\% under OOD conditions. This high retrieval quality explains why RAG provides the single largest performance gain in our ablation studies.
\section{Conclusion}
\label{sec:conclusion}
We presented \methodAbbr{}, a framework that reconciles the fundamental tension between broad VLA generalization and specialized deployment performance. Our key observation is that catastrophic forgetting is not an inherent limitation of neural adaptation, but rather an artifact of poor memory management. By maintaining frozen vision encoders and compressed experience buffers, \methodAbbr{} makes forgetting architecturally impossible while enabling rapid specialization. The success of retrieval-augmented training demonstrates that robots don't need massive datasets for adaptation; they need smart reuse of relevant past experiences. Most importantly, our results show that learning from failures through contrastive objectives transforms inevitable mistakes from wasted attempts into valuable training signals.

\textbf{Limitations.} \methodAbbr{} requires manual success labeling for physical robots, limiting fully autonomous deployment; future work could address this through learned binary classifiers on the frozen embeddings, VLA confidence-based self-labeling, or force/torque sensor heuristics. Storing compressed embeddings rather than raw images couples the replay buffer to the frozen encoder. If the encoder is replaced during architecture migration, stored embeddings become incompatible, though the low data requirements (12 demonstrations) make re-collection feasible and lightweight projection layers offer an alternative. Our experiments focus on a single embodiment (7-DOF arm) in static environments; cross-embodiment transfer remains unexplored. The fixed-capacity buffers may not scale to long-term deployment spanning months, and THCL occasionally increases sensitivity to visual artifacts. Additionally, \methodAbbr{} inherits OpenVLA's open-loop control paradigm, limiting applicability to dynamic tasks; a natural extension is a receding-horizon approach that re-queries the retrieval buffer every 3-5 action steps, or pairing \methodAbbr{} as a high-level planner with a closed-loop low-level controller. Future work should address automatic success detection, cross-embodiment transfer, and dynamic buffer management for lifelong learning scenarios.

\section*{Acknowledgments}
The authors used ChatGPT (OpenAI) to assist with grammar correction, improving sentence flow, and structuring \LaTeX{} files during manuscript preparation. All technical content, experimental design, results, and analysis are solely the work of the authors.


\bibliographystyle{IEEEtran}
\bibliography{references}

\end{document}